\documentclass{article}

\usepackage{microtype}
\usepackage{graphicx}
\usepackage{subcaption}
\usepackage{booktabs} %
\usepackage{xspace}

\usepackage{hyperref}

\usepackage[preprint]{icml2026}

\usepackage{amsmath}
\usepackage{amssymb}
\usepackage{mathtools}
\usepackage{amsthm}
\usepackage{enumitem}
\usepackage{amsfonts}
\usepackage{bm}
\usepackage{bbm}
\usepackage{hhline}
\usepackage{multirow}
\usepackage{gensymb}

\usepackage{enumerate}
\usepackage{enumitem}  %
\usepackage{makecell}
\usepackage{pifont}
\usepackage{titlecaps}

\usepackage[capitalize,noabbrev]{cleveref}

\theoremstyle{plain}

\theoremstyle{definition}

\theoremstyle{remark}

\usepackage[textsize=tiny]{todonotes}

\newcommand{\myparagraph}[1]{\noindent\textbf{#1}}

\renewcommand{\paragraph}[1]{\vspace{0.1cm}\noindent\textbf{#1}}

\definecolor{MyDarkBlue}{rgb}{0,0.08,1}
\definecolor{MyAqua}{rgb}{0,0.7,0.7}
\definecolor{MyDarkGreen}{rgb}{0.02,0.6,0.02}
\definecolor{MyDarkRed}{rgb}{0.8,0.02,0.02}
\definecolor{MyDarkOrange}{rgb}{0.40,0.2,0.02}
\definecolor{MyPurple}{RGB}{111,0,255}
\definecolor{MyRed}{rgb}{1.0,0.0,0.0}
\definecolor{MyGold}{rgb}{0.75,0.6,0.12}
\definecolor{MyDarkgray}{rgb}{0.66, 0.66, 0.66}

\definecolor{Cardinal}{rgb}{0.549,0.082,0.082}

\newcommand{\ie}{\textit{i.e.}}

\newcommand{\hquad}{\;\;}

\newcommand{\model}{SurfPhase\xspace}

\newcommand{\titleshort}{SurfPhase: 3D Interfacial Dynamics in Two-Phase Flows from Sparse Videos}

\newcommand{\titlefull}{SurfPhase: 3D Interfacial Dynamics in Two-Phase Flows from Sparse Videos}

\newcommand{\keywords}{3D Interfacial Dynamics, Two-phase Flow, Sparse View 3D Reconstruction, Dynamic 3D Reconstruction}

\icmltitlerunning{\titleshort} %

\begin{document}

\twocolumn[
  \icmltitle{\titlefull}

  \icmlsetsymbol{equal}{*}

  \begin{icmlauthorlist}
    \icmlauthor{Yue Gao}{equal,stanford} \hquad
    \icmlauthor{Hong-Xing Yu}{equal,stanford} \hquad
    \icmlauthor{Sanghyeon Chang}{uci} \hquad
    \icmlauthor{Qianxi Fu}{uci}\\
    \icmlauthor{Bo Zhu}{gatech} \hquad
    \icmlauthor{Yoonjin Won}{uci} \hquad
    \icmlauthor{Juan Carlos Niebles}{stanford} \hquad
    \icmlauthor{Jiajun Wu}{stanford}
  \end{icmlauthorlist}

  \icmlaffiliation{stanford}{Computer Science Department, Stanford University} %
  \icmlaffiliation{uci}{Mechanical and Aerospace Engineering, University of California, Irvine} %
  \icmlaffiliation{gatech}{School of Interactive Computing, Georgia Institute of Technology} %

  \icmlcorrespondingauthor{Yue Gao}{\mbox{yuegao}@cs.stanford.edu}

  \icmlkeywords{\keywords}

  \vskip 0.3in
]

\printAffiliationsAndNotice{\icmlEqualContribution}

\begin{abstract}
Interfacial dynamics in two-phase flows govern momentum, heat, and mass transfer, yet remain difficult to measure experimentally. Classical techniques face intrinsic limitations near moving interfaces, while existing neural rendering methods target single-phase flows with diffuse boundaries and cannot handle sharp, deformable liquid-vapor interfaces. We propose \model, a novel model for reconstructing 3D interfacial dynamics from sparse camera views. Our approach integrates dynamic Gaussian surfels with a signed distance function formulation for geometric consistency, and leverages a video diffusion model to synthesize novel-view videos to refine reconstruction from sparse observations. We evaluate on a new dataset of high-speed pool boiling videos, demonstrating high-quality view synthesis and velocity estimation from only two camera views.  Project website: \url{https://yuegao.me/SurfPhase}.

\end{abstract}

\section{Introduction}

Multiphase flows, such as boiling and condensation, are governed by interfacial physics in which momentum, heat, and mass transfer are strongly coupled through a dynamically evolving interface~\citep{kharangate2017review}. The interface is deformable, continuously changing topology, and must be treated as a moving control surface rather than a static boundary. Interfacial velocity is a fundamental variable: it directly determines momentum transport, shear stress, and phase slip, which in turn govern heat transfer and interfacial stability. Yet despite its central role, interfacial velocity remains difficult to access experimentally, and many studies rely on geometric observations or global metrics alone~\citep{o2020review}, while recent vision-based pipelines have begun digitizing interface geometry and dynamics directly from videos~\citep{chang2023bubblemask, chang2025eventflow, suh2024vision}.

\begin{figure}[t]
  \centering
  \includegraphics[width=0.95\linewidth]{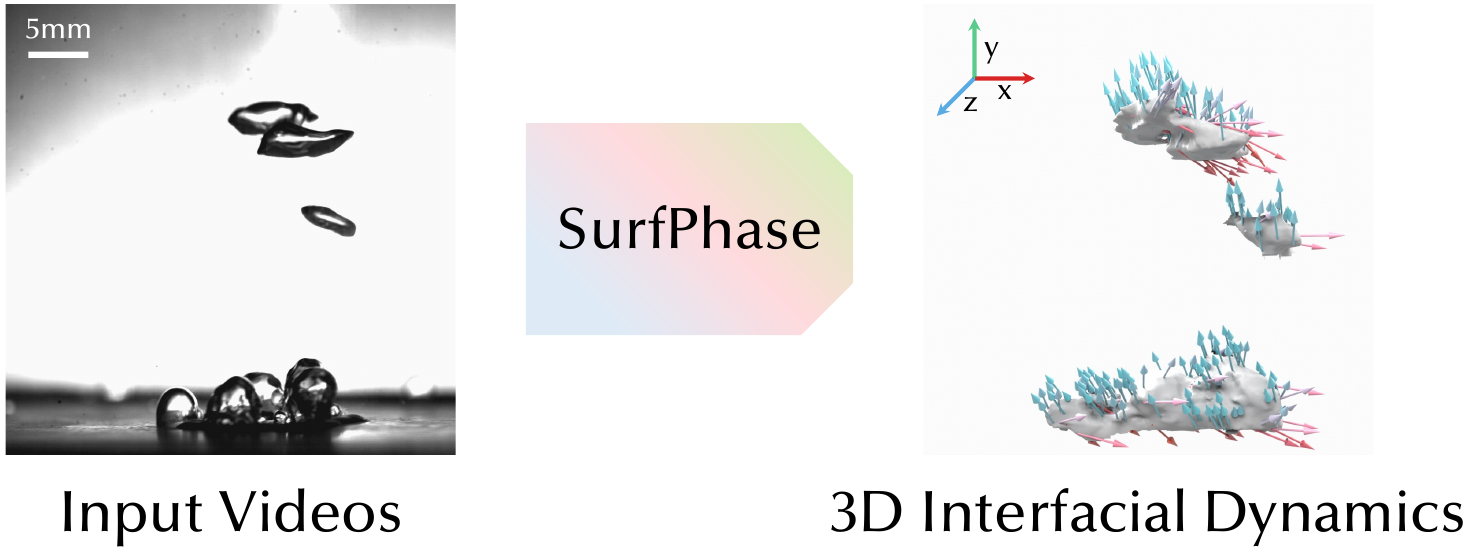}
  \caption{We introduce a new task: Reconstructing 3D interfacial dynamics from sparse-view videos.}
  \vspace{-3mm}
  \label{fig:teaser} 
\end{figure}

Classical measurement techniques face intrinsic limitations near moving interfaces: intrusive probes disturb flow, while optical methods such as PIV suffer from tracer exclusion and occlusion~\citep{kahler2012uncertainty,deen2002two}. Computational fluid dynamics provides detailed three-dimensional fields, but requires inputs difficult to obtain independently (boundary conditions, nucleation site distributions, semi-empirical closure models), and without direct experimental measurements of the predicted quantities, simulation accuracy cannot be independently verified. Recently, computer vision and neural rendering offer an opportunity to bridge this gap to enable non-intrusive three-dimensional reconstruction and velocity estimation of evolving interfaces directly from videos. This holds promise for grounding computational models in real observations while providing access to interfacial kinematics at a level of detail previously available only in simulation.

Recent advances in neural rendering and 3D reconstruction have enabled recovery of dynamic scenes from videos~\citep{chu2022physics,yu2024inferring,franz2021global}. Methods based on neural radiance fields and Gaussian splatting achieve photorealistic novel-view synthesis for static and dynamic scenes. For fluids, FluidNexus~\citep{gao2025fluidnexus} demonstrated reconstruction and prediction of single-phase smoke by combining physics-based simulation with neural rendering. However, these methods target gaseous flows with diffuse boundaries and rely on simulation priors that do not transfer to two-phase systems. The sharp, deformable liquid-vapor interfaces in boiling and condensation undergo topological changes, exhibit specular reflections and refractions, and require geometric accuracy for meaningful velocity estimation. Therefore, existing approaches cannot address these requirements.

We propose \model, a novel approach to reconstructing 3D interfacial dynamics in two-phase flows from sparse camera observations (Figure~\ref{fig:teaser}). Unlike general scenes where dense camera arrays are feasible, two-phase flow experiments impose severe constraints on optical access due to thermal conditions and equipment placement, often limiting practical setups to as few as two synchronized views. Our method represents the scene using dynamic Gaussian surfels: oriented surface elements that explicitly model interface geometry over time. To capture liquid-vapor interfaces, we integrate a signed distance function (SDF) formulation that enforces geometric consistency and encourages coherent surface reconstruction. To compensate for limited viewing angles, we introduce a generative refinement strategy: a video diffusion model trained on single-view footage learns visual priors of interface appearance and motion, and then synthesizes plausible novel viewpoints incorporated for a refined reconstruction. From tracked 3D interface surfaces, we can extract meshes and estimate velocity profiles in calibrated metric coordinates.

Our contributions are threefold:
\begin{itemize}[leftmargin=*, itemsep=1pt, topsep=2pt, parsep=0pt, partopsep=0pt]
    \item We introduce the problem of reconstructing 3D interfacial dynamics of two-phase flows from sparse-view (as few as two views) videos, establishing a new task at the intersection of computer vision and multiphase flow research.
    \item We propose \model, which integrates dynamic Gaussian surfels with SDF constraints and diffusion-based novel-view video synthesis to achieve interface reconstruction from only two views.
    \item We collect a dataset of high-speed pool boiling videos comprising 200 monocular videos and a synchronized dual-view video pair with physically calibrated metric scale, allowing training and evaluation for the new task of 3D interfacial reconstruction.
\end{itemize}

\section{Related Works}

\myparagraph{Video-based fluid analysis.}
Fluid flow measurement from visual observations has been extensively studied in experimental fluid mechanics. Classical approaches include active sensing methods such as laser scanners~\citep{hawkins2005acquisition} and structured light~\citep{gu2012compressive}, as well as passive techniques like particle image velocimetry (PIV)\citep{adrian2011particle,elsinga2006tomographic}. To enable broader applicability beyond controlled laboratory settings, recent methods have explored video-based fluid analysis using tomographic reconstruction\citep{gregson2014capture,okabe2015fluid,zang2020tomofluid} or neural and differentiable rendering~\citep{franz2021global}. These approaches typically require multi-view synchronized videos and are often aided by differentiable physics simulation to constrain the solution space. Physics-informed methods such as Physics-Informed Neural Fluid~\citep{chu2022physics} and HyFluid~\citep{yu2024inferring} incorporate physical priors to maintain plausibility. Most recently, FluidNexus~\citep{gao2025fluidnexus} demonstrated reconstruction and prediction of single-phase smoke flows from a single video by combining physics-based simulation with video generation. However, these methods target gaseous flows with diffuse boundaries and rely on simulation priors that do not transfer to two-phase systems. Our work addresses this gap by enabling 3D interfacial reconstruction from sparse observations, as few as two views, without relying on fluid simulation priors.

\myparagraph{Dynamic 3D reconstruction.}
Reconstructing dynamic 3D scenes from videos remains a fundamental challenge in computer vision~\citep{bansal20204d}. Recent progress has been driven by representing scenes with dynamic radiance fields, including dynamic NeRFs~\citep{nerf,li2021neural,park2021hypernerf,pumarola2021d,fridovich2023k,gao2021dynamic} and dynamic 3D Gaussians~\citep{lin2024gaussian,luiten2024dynamic,li2024spacetime}, optimized through differentiable rendering from multi-view observations. By leveraging monocular depth estimation~\citep{ranftl2022midas,ke2023repurposing}, recent methods have extended these capabilities to monocular video input~\citep{som2024,lei2024mosca,zhang2024monst3r}. For surface reconstruction specifically, 2D Gaussian Splatting~\citep{huang20242d} introduced Gaussian surfels that collapse 3D Gaussians onto oriented disks, enabling geometrically accurate radiance field reconstruction. However, these methods assume sufficient multi-view coverage, which is often unattainable in two-phase flow experiments due to thermal and physical constraints on camera placement. Pretrained depth priors also fail to generalize to two-phase imagery with specular interfaces and refractive effects. Our approach addresses this limitation by incorporating a video diffusion model to synthesize plausible novel viewpoints, providing additional supervision that compensates for limited camera coverage.

\myparagraph{Video generation.}
Video generation has advanced rapidly with the development of diffusion models~\citep{hong2022cogvideo,alignyourlatents2023,svdblattmann2023,emuvideo2023,lumiere2024,sora2024}. Recent models such as Sora~\citep{sora2024} have demonstrated remarkable capability in synthesizing realistic physical phenomena, including fluid dynamics and object interactions. This has motivated the use of video generation as a prior for 3D reconstruction and physical scene understanding. Several methods leverage video diffusion models to synthesize novel views for static scene reconstruction~\citep{he2024cameractrl,ren2025gen3c}, while others use generated videos to provide temporal supervision for dynamic scenes~\citep{gao2025fluidnexus}. However, these approaches have focused on general objects or single-phase flows, and the applicability of video generation priors to two-phase systems with sharp interfaces remains unexplored. In our work, we train a video diffusion model on our collected single-view pool boiling footage to learn visual priors specific to interface appearance and motion, providing additional constraints that refine interface geometry reconstruction.

\begin{figure*}[t]
  \centering
  \includegraphics[width=0.99\linewidth]{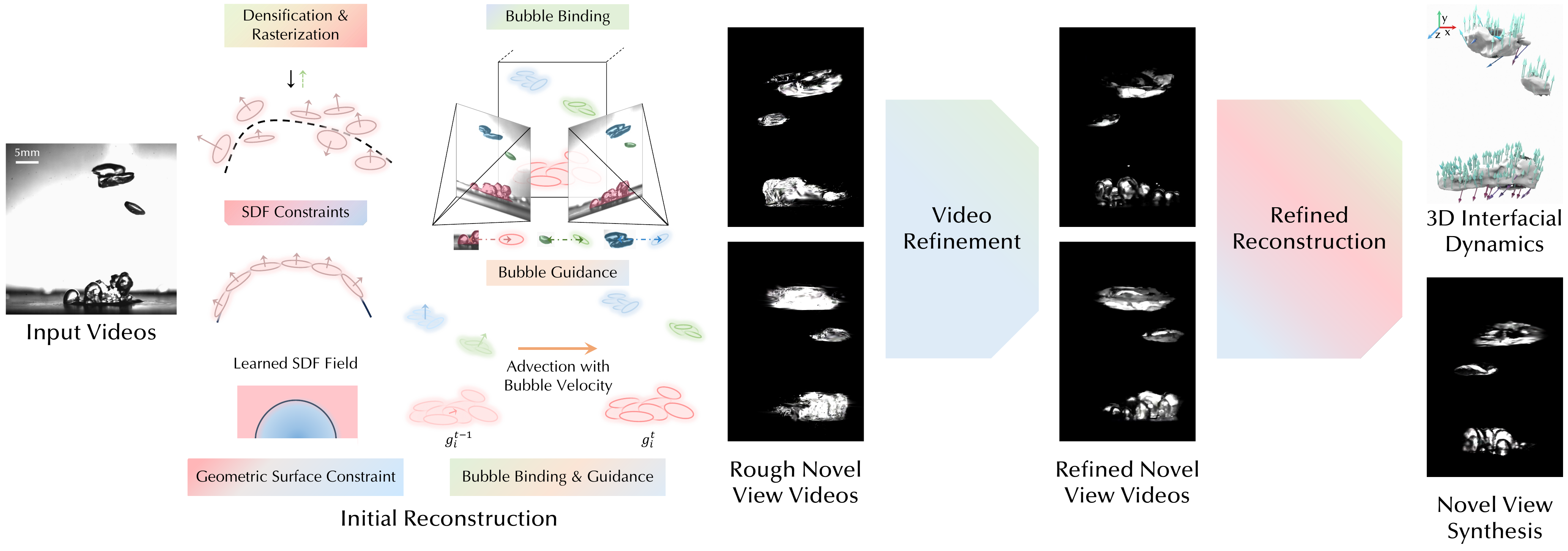}
  \vspace{-1.5mm}
  \caption{\textbf{Overview.} Given input videos of two-phase flows, \model\ reconstructs the 3D appearance, geometry, and velocity. During the refined reconstruction stage, we use same geometric surface constraint and bubble guidance as in initial reconstruction. We summarize the pipeline in Alg.~\ref{algo:dyn_bubble_velocity} in the appendix.}
  \label{fig:overview} 
  \vspace{-3mm}
\end{figure*}

\section{Approach}

\myparagraph{Problem statement.}
Given $C$ synchronized videos $\{\mathcal{V}^c\}_{c=1}^{C}$, where $\mathcal{V}^c = (I_1^c, \ldots, I_T^c)$ contains $T$ frames $I_t^c \in \mathbb{R}^{H \times W \times 3}$ captured from a calibrated camera with projection matrix $\pi_c \in \mathbb{R}^{3 \times 4}$, we aim to reconstruct the 3D geometry, appearance, and velocity of the liquid-vapor interface over time. Due to thermal and physical constraints on camera placement in two-phase flow experiments, $C$ is typically very small (as few as two views).

\myparagraph{Challenges.}
This task presents two major challenges. First, reconstructing dynamic 3D interfaces from sparse-view videos is severely ill-posed: the liquid-vapor interface undergoes rapid deformation and topological changes, and standard multi-view constraints become insufficient when only a few camera views are available. Second, estimating interfacial velocity requires not only per-frame geometry but also point-level 3D tracking of the interface over time. Differentiable rendering-based optimization alone does not provide such correspondences~\citep{xing2022differentiable,gao2025fluidnexus}, as it optimizes appearance agreement without explicitly enforcing temporal consistency of individual surface elements.

\myparagraph{Overview.}
We propose \model\ to address these challenges (Figure~\ref{fig:overview}). For the first challenge, we integrate geometric surface constraints with learned priors of visual dynamics through a two-stage reconstruction pipeline. In the initial reconstruction stage (left of Figure~\ref{fig:overview}), we represent the scene using dynamic Gaussian surfels augmented with a signed distance function (SDF) formulation that enforces geometric consistency. We optimize this representation from the sparse input views and render videos from multiple novel viewpoints. In the refinement stage (right of Figure~\ref{fig:overview}), we leverage a video diffusion model trained on single-view two-phase flow footage to refine these rough novel-view renderings, producing temporally coherent videos that capture plausible interface appearance and motion. We then incorporate the refined videos as additional supervision for a second round of reconstruction, yielding improved interface geometry. For the second challenge, we introduce bubble-guided velocity estimation (middle-left of Figure~\ref{fig:overview}): we bind each Gaussian surfel to an individual bubble instance using multi-view segmentation, then estimate each bubble's 3D velocity to initialize surfel positions across frames, enabling coherent tracking and metric velocity estimation. We show an algorithm in Alg.~\ref{algo:dyn_bubble_velocity} in the appendix.

\subsection{Two-Stage Reconstruction}

In the initial reconstruction stage, we aim to recover 3D appearance and geometry from sparse input views in order to synthesize novel-view videos for subsequent refinement. We represent the dynamic scene using Gaussian surfels augmented with signed distance function (SDF) values, and optimize this representation through a combination of appearance and geometry losses.

\myparagraph{Scene representation.}
We represent the liquid-vapor interface at each timestep $t$ using a set of Gaussian surfels $\{g_i\}_{i=1}^{N_t}$, where each surfel is characterized by its attributes $\{\boldsymbol{\mu}_i, \mathbf{s}_i, \mathbf{r}_i, \mathbf{c}_i, f_i\}$, representing position, scale, rotation, color, and SDF value, respectively. Following 2D Gaussian Splatting~\citep{huang20242d}, each surfel is a flat disk defined in a local tangent plane with two principal axes $\mathbf{t}_u$ and $\mathbf{t}_v$, and normal direction $\mathbf{n}_i = \mathbf{t}_u \times \mathbf{t}_v$. The surfels are rendered to images via differentiable rasterization. Unlike standard Gaussian splatting that directly optimizes opacity $o_i$, we derive opacity from the SDF value through a bell-shaped transformation, as described below.

\myparagraph{Geometric surface constraint.}
Standard Gaussian surfels are appearance-oriented and do not explicitly model surface geometry, making them ill-suited for interface reconstruction where geometric accuracy is essential. Inspired by recent work on SDF-based reconstruction~\citep{zhu_2025_dsdf}, we adopt a discretized signed distance function formulation. Each surfel stores a learnable SDF value $f_i$, from which opacity is analytically derived via a logistic transformation:
\begin{equation}
o_i = \mathcal{T}_\gamma(f_i) = \frac{4 e^{-\gamma f_i}}{(1 + e^{-\gamma f_i})^2},
\end{equation}
where the global parameter $\gamma$ controls the sharpness of the opacity transition. To ensure surfels tightly approximate the interface, we employ a median-guided regularization that dynamically adjusts $\gamma$. Specifically, we compute a target sharpness $\gamma_m = -\log(3 - 2\sqrt{2}) / |f|_m$ based on the median absolute SDF value $|f|_m$ across all surfels, and apply a hinge loss $\mathcal{L}_\gamma = \max(\gamma_m - \gamma, 0)$ to progressively narrow the transition region during optimization.

To encourage the SDF field to satisfy the Eikonal property without ground-truth supervision, we use a projection-based consistency loss. Since the normal direction of each surfel provides the gradient direction $\mathbf{n}_i = \nabla f_i / |\nabla f_i|$, we project each surfel onto the estimated zero-level set via $\boldsymbol{\mu}_i^{\text{proj}} = \boldsymbol{\mu}_i - f_i \mathbf{n}_i$. We then minimize the discrepancy between the depth of this projected point and the alpha-blended rendered depth:
\vspace{-1mm}
\begin{equation}
\mathcal{L}_p = \frac{1}{N} \sum_{i=1}^{N} \mathbf{1}[\epsilon_i \leq \varepsilon] \cdot \epsilon_i, \quad \epsilon_i = |D_{\text{render}} - D_{\text{proj}}^i|,
\end{equation}
where $\varepsilon=0.1$ is a threshold that excludes outliers arising from occlusions or self-intersections.

\myparagraph{Optimization.}
We formulate dynamic reconstruction as sequential per-frame optimization from $t = 1$ to $T$. At each frame, we minimize a combined loss:
\vspace{-1mm}
\begin{equation}
\mathcal{L} = \lambda_{\text{app}} \mathcal{L}_{\text{app}} + \lambda_{\text{geo}} \mathcal{L}_{\text{geo}}.
\end{equation}
The appearance loss measures the difference between rendered images $\hat{I}_t^c$ and reference frames $I_t^c$ across all input views:
\vspace{-1mm}
\begin{equation}
\mathcal{L}_{\text{app}} = \sum_{c=1}^{C} \mathcal{L}_1(I_t^c, \hat{I}_t^c) + \mathcal{L}_{\text{SSIM}}(I_t^c, \hat{I}_t^c).
\end{equation}
The geometry loss combines normal consistency, the SDF median regularization, and the depth projection loss:
\begin{equation}
\mathcal{L}_{\text{geo}} = \mathcal{L}_n + \mathcal{L}_\gamma + \mathcal{L}_p,
\end{equation}
where $\mathcal{L}_n$ encourages rendered normals to be consistent with the normals derived from depth. At the first frame ($t = 0$), we allow densification and pruning of surfels; for subsequent frames, we disable densification to maintain temporal correspondence.

\myparagraph{Video refinement.}
After initial reconstruction, the optimized surfels can render reasonable views close to the input cameras but exhibit artifacts at distant viewpoints due to insufficient multi-view constraints. To address this, we leverage a video diffusion model to refine novel-view renderings. We first create $S$ novel camera viewpoints $\{\pi_s\}_{s=1}^{S}$ orbiting the scene and render videos $\{\tilde{\mathcal{V}}^s\}_{s=1}^{S}$ from these viewpoints. These rough videos capture basic interface structure but lack realistic appearance and motion dynamics. We then apply SDEdit-based refinement~\citep{meng2021sdedit,gao2025fluidnexus} using a video diffusion model $\mathcal{G}$~\citep{wan2025} fine-tuned on our collected single-view two-phase flow videos:
\begin{equation}
\hat{\mathcal{V}}^s = \mathcal{G}(\tilde{\mathcal{V}}^s; \lambda_{\text{refine}}),
\end{equation}
where $\lambda_{\text{refine}}$ controls the strength of generative refinement. The refined videos $\{\hat{\mathcal{V}}^s\}_{s=1}^{S}$ exhibit temporally coherent interface appearance and physically plausible deformation learned from the training data. We incorporate these refined videos as additional supervision in a second optimization stage, using adaptive loss weights that assign higher weight to input views and lower weight to distant novel views to account for uncertainty in the generated content.

\subsection{Bubble-Guided Velocity Estimation}

Estimating interfacial velocity from the reconstructed scene requires not only per-frame geometry but also surfel-wise 3D tracking of the interface over time. However, differentiable rendering-based optimization alone does not establish such correspondences: it minimizes appearance error without explicitly enforcing that individual surfels track consistent surface points across frames. In practice, gradients with respect to surfel positions are often ineffective, causing surfels to remain nearly stationary while adjusting their scale and color to fit the target appearance, rather than translating to follow the moving interface~\citep{xing2022differentiable}. To address this challenge, our key idea is to bind each Gaussian surfel to an individual bubble instance, then estimate each bubble's 3D velocity to guide the motion of all surfels bound to it.

\myparagraph{Bubble binding.}
We leverage the Segment Anything Model (SAM)~\citep{Kirillov_2023_ICCV} to obtain 2D instance segmentation masks for each bubble across all input views. For each view $c$, we extract bubble masks $\mathcal{M}^c \in \{0,1\}^{T \times B \times H \times W}$, where $B$ is the maximum number of bubble instances. To associate masks across views, we assume that bubbles with similar vertical positions in image space correspond to the same 3D instance. Since 2D tracking can fail due to occlusions or bubble merging and splitting, we aggregate information from all input views to improve robustness. Specifically, we traverse viewpoints in order of decreasing number of valid bubble masks (where validity is determined by a minimum area threshold). For each surfel $g_i$, we project its position $\boldsymbol{\mu}_i$ to viewpoint $c$ and assign bubble index $b$ if the projection falls within mask $\mathcal{M}^c_b$. Surfels already assigned in a previous viewpoint are skipped. This strategy prioritizes views with more complete bubble visibility, allowing us to resolve ambiguities caused by occlusion in individual views.

\myparagraph{Bubble velocity-guided initialization.}
Given bubble assignments, we estimate each bubble's 3D velocity from the previous frame and use it to initialize surfel positions for the current frame. For bubble $b$ at frame $t-1$, we compute the weighted centroid:
\begin{equation}
\mathbf{c}_{b,t-1} = \frac{\sum_{i: b_i = b} w_i \boldsymbol{\mu}_{i,t-1}}{\sum_{i: b_i = b} w_i}, \quad w_i = \sigma(o_i) \cdot s_{i,x} \cdot s_{i,y},
\end{equation}
where $\sigma(\cdot)$ is the sigmoid function, $o_i$ is the opacity, and $s_{i,x}, s_{i,y}$ are the surfel's scale components. The bubble velocity is then estimated by finite difference between consecutive centroids:
\begin{equation}
\mathbf{u}_b = \frac{\mathbf{c}_{b,t-1} - \mathbf{c}_{b,t-2}}{\Delta t}.
\end{equation}
We initialize all surfels bound to bubble $b$ at frame $t$ by advecting their positions from the previous frame:
\begin{equation}
\boldsymbol{\mu}_{i,t}^{\text{init}} = \boldsymbol{\mu}_{i,t-1} + \Delta t \cdot \mathbf{u}_{b_i}.
\label{eq:advection}
\end{equation}
This initialization provides a physically informed starting point for optimization, encouraging surfels to track coherent surface regions rather than remaining stationary and deforming to match appearance. For the nucleation region at the heated surface (bubble index $0$), we instead use the per-surfel velocity from the previous frame to accommodate more complex local motion.

\myparagraph{Velocity estimation.}
Given the tracked surfel positions across frames, we estimate interfacial velocity by finite difference. The velocity of each surfel $g_i$ at frame $t$ is:
\begin{equation}
\mathbf{v}_{i,t} = \frac{\boldsymbol{\mu}_{i,t} - \boldsymbol{\mu}_{i,t-1}}{\Delta t}.
\end{equation}
For bubble-level velocity, we compute the centroid velocity:
\begin{equation}
\mathbf{v}_{b,t} = \frac{\mathbf{c}_{b,t} - \mathbf{c}_{b,t-1}}{\Delta t}.
\end{equation}
Since the cameras are calibrated in metric coordinates, these velocities are directly expressed in physical units (m/s), enabling quantitative comparison with independent measurements.

\section{Experiments}

\myparagraph{Datasets.}
We collect a new dataset of high-speed pool boiling videos for training and evaluation. For training the video diffusion model, we capture $200$ monocular videos at $2{,}000$ FPS using high-speed cameras with shallow depth of field. These videos are not calibrated and serve as a source for learning visual priors of liquid-vapor interfacial dynamics. For evaluation, we capture a separate synchronized dual-view video pair using two high-speed cameras positioned at a viewing angle of $35\degree$. Each video contains at least $1{,}500$ frames with matched lighting conditions across views to ensure appearance consistency. We carefully calibrate the camera intrinsic and extrinsic parameters to enable metric-space reconstruction and velocity estimation.

For quantitative evaluation with ground-truth geometry and velocity, we construct a synthetic two-phase flow scene using Houdini. We simulate vapor generation using the POP Source operator emitting particles from a heated surface, with FLIP Solver handling liquid-gas interaction and container collision. The particles are rendered as randomly deforming spheres to mimic interface dynamics. We render the scene from three viewpoints: two side views serve as input and the middle view provides ground truth for novel view synthesis evaluation.

\begin{figure}[t]
    \centering
    \includegraphics[width=0.89\linewidth]{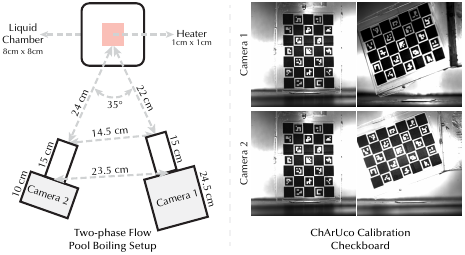}
    \caption{Our real two-phase flow pool boiling setup (left) and samples of our calibration image pairs (right). 
    }
    \vspace{-4mm}
    \label{fig:calibration}
\end{figure}

\myparagraph{Real data calibration.}
Recent feed-forward models for camera pose estimation~\citep{wang2025vggt,cut3r} struggle with our data due to the large domain gap from natural images. Structure-from-motion methods such as COLMAP~\citep{colmap} estimate only relative camera parameters without metric scale alignment. Furthermore, high-speed cameras are sensitive to lighting conditions, causing appearance variations across views that degrade feature matching accuracy. The sparse camera placement required by thermal constraints in two-phase flow experiments further limits the applicability of SfM-based calibration.

We therefore adopt ChArUco board calibration for accurate metric-space alignment. We print a small transparent plastic ChArUco board ($5 \times 7$ grid, $3\text{mm}$ markers, $4\text{mm}$ squares) and place it at the center of the experimental setup. With the two high-speed cameras fixed, we capture multiple images of the board at varying positions and orientations. We then use the standard ChArUco calibration routines from OpenCV~\citep{opencv_library} to estimate camera intrinsics and extrinsics. We show an illustration in Fig.~\ref{fig:calibration}.

\myparagraph{Baselines.}
Since our work introduces a new task, no existing method directly addresses 3D interfacial reconstruction in two-phase flows from sparse views. We compare against the most relevant prior work: FluidNexus~\citep{gao2025fluidnexus}, which reconstructs single-phase smoke flows and estimates velocity fields but is not designed for two-phase systems with sharp interfaces. For 3D appearance reconstruction, we include 4DGS~\citep{yang2023gs4d}, a state-of-the-art dynamic scene reconstruction method based on temporal deformation. For geometry evaluation, we compare against 2DGS~\citep{huang20242d}, the Gaussian surfel representation on which our method builds. For velocity estimation, we also implement \citet{zhang2026multi} which combines a metric depth predictor~\citep{video_depth_anything} with an optical flow estimator~\citep{wang2025waft} to obtain metric-space point clouds and motion.

\myparagraph{Task settings.}
We reconstruct 3D appearance, geometry, and velocity of the liquid-vapor interface. Thus, we evaluate \model\ on three tasks: (1) novel view video synthesis, (2) interfacial geometry reconstruction, and (3) 3D velocity estimation. For velocity evaluation, we quantitatively evaluate bubble-level velocity, and visualize interfacial velocity fields for qualitative assessment.

\myparagraph{Metrics.}
For novel view synthesis, we compute standard metrics including L1, PSNR, and SSIM on synthetic data where ground-truth novel views are available. For interfacial geometry reconstruction, we report Chamfer distance on synthetic data. For 3D bubble velocity estimation, we compute mean L1 error on both real and synthetic data. On real data where ground-truth 3D bubble velocity is unavailable, we measure velocity along the $x$-axis (horizontally aligned with the pixel coordinate) and $y$-axis (aligned with gravity) using a physical calibration checkerboard to establish pixel-to-metric correspondence, and report relative error against this independent measurement. All metrics are computed per frame and averaged over the video sequence.

\myparagraph{Implementation details.}
For Gaussian surfel optimization, we follow 2DGS~\citep{huang20242d} for learning rates of position, scale, rotation, and color attributes, and set the SDF value learning rate to $0.05$. We use the Adam optimizer~\citep{kingma2014adam} with $\epsilon = 10^{-15}$. Initial velocities are set to $\mathbf{u}_{b=0} = [0.03, 0.03, 0]^\top$ for the nucleation region and $\mathbf{u}_{b>0} = [0.07, 0.3, 0]^\top$ for rising interfaces.

After initial reconstruction, we render $S{=}11$ rough novel-view videos by orbiting the scene center at $30\degree$ increments starting from Camera 1. For video refinement, we fine-tune the pretrained Wan2.1~\citep{wan2025} text-to-video model using LoRA~\citep{hu2021lora} with rank $32$ on our $200$ monocular videos. We use AdamW~\citep{loshchilov2017decoupled} with learning rate $10^{-4}$ and train for $10$ epochs on $T = 101$ consecutive frames per video. Text captions are generated using ChatGPT from the first frame of each video. We set refinement strength $\lambda_{\text{refine}} = 0.2$ for novel views $s \in \{1, 2, 3, 9, 10, 11\}$ close to the input cameras and $\lambda_{\text{refine}} = 0.35$ for more distant views $s \in [4, 8]$.

During the refinement stage, we incorporate refined novel-view videos with adaptive loss weights: input views receive weight $w_c = 1.0$, while novel views receive $w_s = 0.5$ for $s \in \{1, 2, 3, 9, 10, 11\}$ and $w_s = 0.25$ for $s \in [4, 8]$, accounting for higher uncertainty in distant synthesized views.

\begin{table}[t]
    \centering
    \caption{Quantitative evaluation. ``CD'' denotes Chamfer distance.}
    \label{tab:numbers}
    \resizebox{\linewidth}{!}{%
    \begin{tabular}{lccc|c|cc}
        \toprule
        \multirow{2}{*}{Method} & \multicolumn{3}{c|}{Novel view video synthesis} & \multicolumn{1}{c|}{Mesh} & \multicolumn{2}{c}{Velocity} \\
                               & L1 $\downarrow$  & PSNR $\uparrow$   & SSIM $\uparrow$    &    CD $\downarrow$   & Synthetic $\downarrow$ & Real $\downarrow$ \\
        \midrule
        \citet{zhang2026multi}                         & -      & -     & -     & -    & 0.115 & 1.82 \\
        2DGS~\cite{huang20242d}            & 0.0021 & 31.30 & 0.989 & 0.44 & -     & -    \\
        4DGS~\cite{Wu_2024_CVPR}           & 0.0016 & 32.56 & 0.992 & 0.20 & -     & -    \\
        FluidNexus~\cite{gao2025fluidnexus}& 0.0022 & 31.37 & 0.985 & 0.49 & 0.248 & 0.371 \\
        \model~(Ours)                      & \textbf{0.0010} & \textbf{35.47} & \textbf{0.994} & \textbf{0.06} & \textbf{0.057} & \textbf{0.013} \\
        \bottomrule
    \end{tabular}}
\end{table}

\begin{figure}[t]
    \centering
    \includegraphics[width=0.9\linewidth]{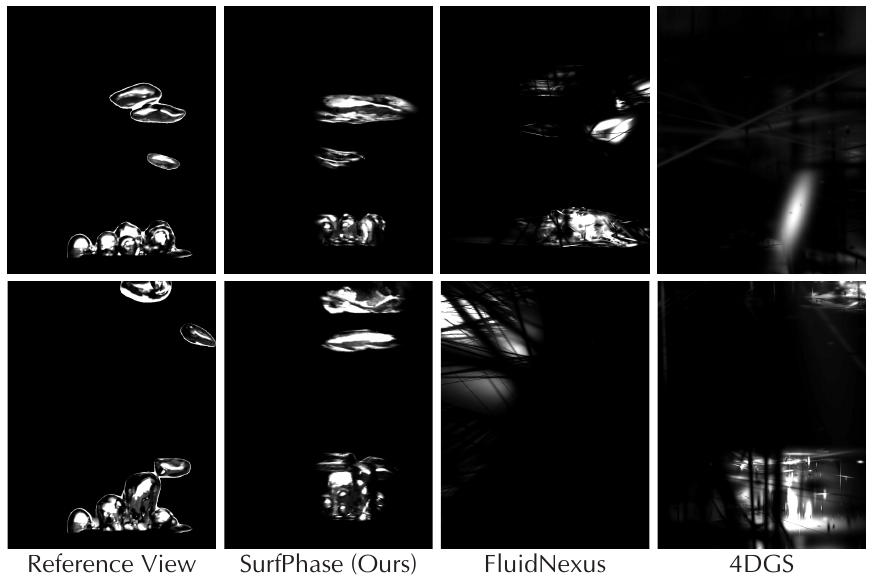}
    \vspace{-1.5mm}
    \caption{Novel view video synthesis on captured real data.
    }
    \vspace{-4mm}
    \label{fig:real_nvs}
\end{figure}

\begin{figure}[t]
    \centering
    \includegraphics[width=0.9\linewidth]{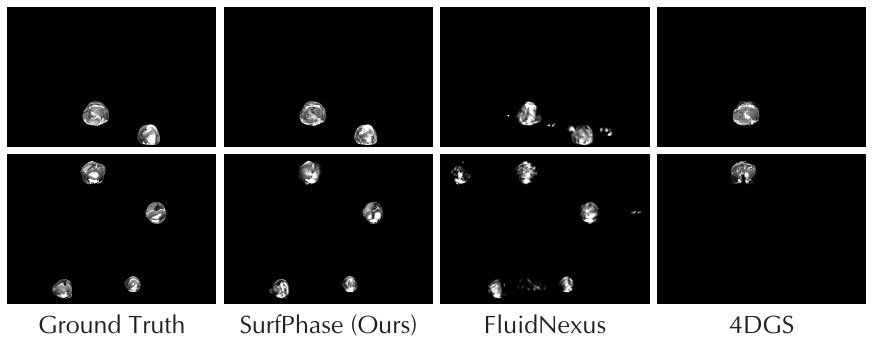}
    \vspace{-1.5mm}
    \caption{Novel view video synthesis on synthetic data.
    }
    \vspace{-3mm}
    \label{fig:syn_nvs}
\end{figure}

\begin{figure}[t]
    \centering
    \includegraphics[width=0.93\linewidth]{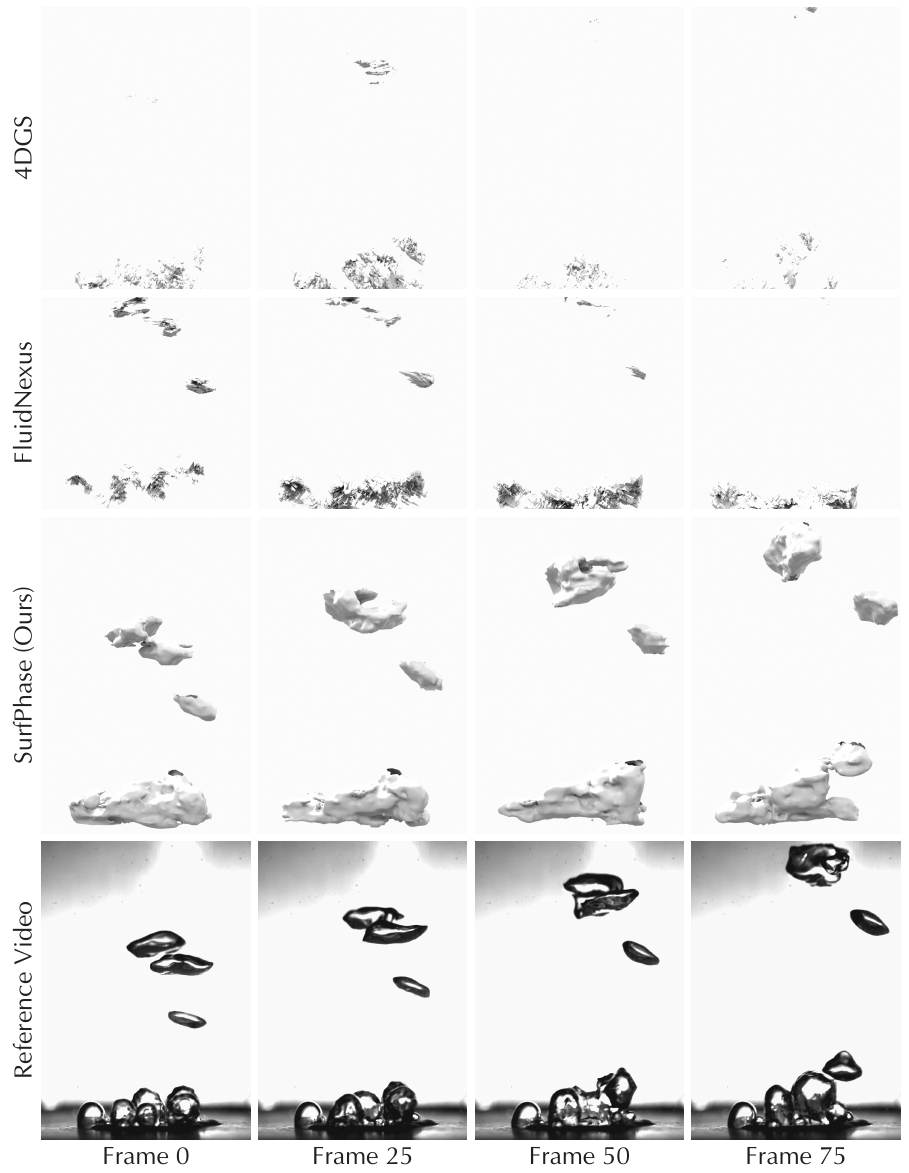}
    \vspace{-1.5mm}
    \caption{Interface reconstruction on captured real data
    }
    \vspace{-3mm}
    \label{fig:real_mesh_static}
\end{figure}

\begin{figure}[t]
    \centering
    \includegraphics[width=0.9\linewidth]{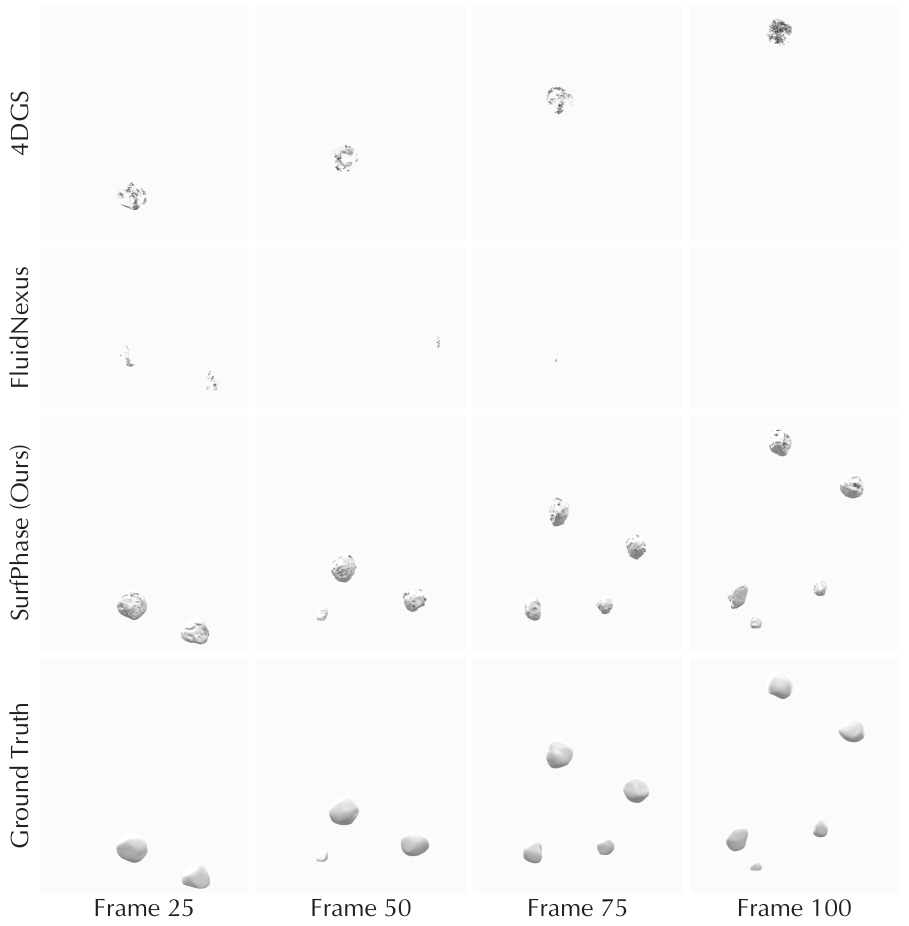}
    \vspace{-1.5mm}
    \caption{Interface reconstruction on synthetic data 
    }
    \vspace{-6mm}
    \label{fig:syn_mesh_static}
\end{figure}

\begin{figure*}
    \centering
    \includegraphics[width=0.9\linewidth]{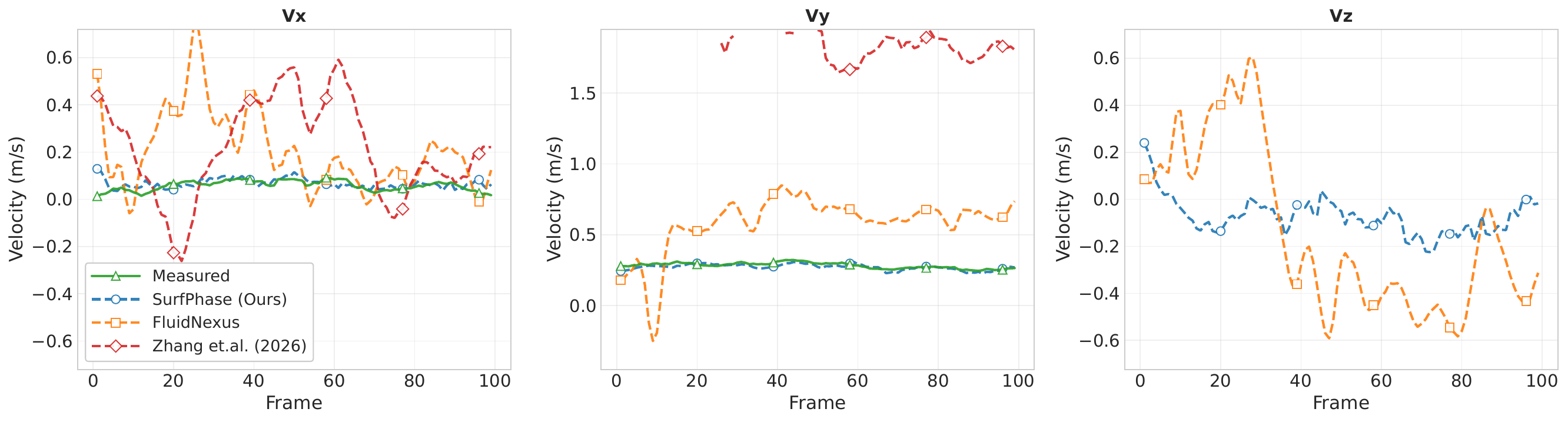}
    \vspace{-1.5mm}
    \caption{Velocity estimation on captured real data. $v_z$ of \citet{zhang2026multi} is too large ($1,000$ times larger than other methods) to show.
    }
    \vspace{-3mm}
    \label{fig:real_velocity}
\end{figure*}

\begin{figure}[t]
    \centering
    \includegraphics[width=0.91\linewidth]{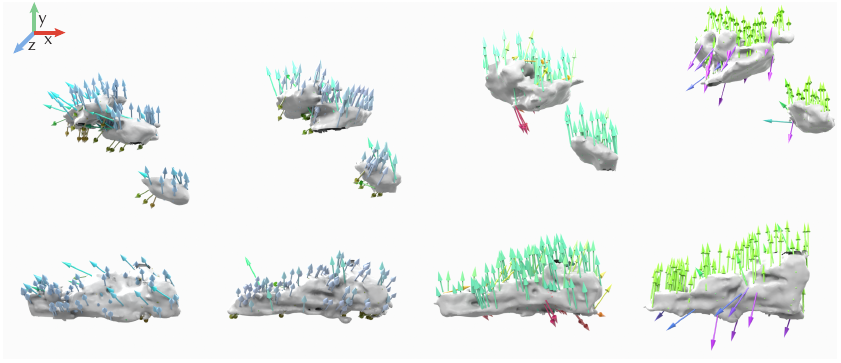}
    \vspace{-1.5mm}
    \caption{3D interfacial velocity visualization across frames. 
    }
    \vspace{-2.9mm}
    \label{fig:real_mesh_velocity}
\end{figure}

\subsection{Comparison to Baselines}

We compare \model\ against baselines on our tasks and report quantitative results in Tab.~\ref{tab:numbers}.

\myparagraph{Novel view video synthesis.}
We show qualitative results on real data in Fig.~\ref{fig:real_nvs} and on synthetic data in Fig.~\ref{fig:syn_nvs}. From the visualization and numbers in Tab.~\ref{tab:numbers}, we observe that \model\ consistently outperforms all baselines. Without dense multi-view constraints, prior 3D reconstruction methods fail to produce reasonable novel-view renderings for two-phase flow scenes. Among them, the best baseline, FluidNexus~\citep{gao2025fluidnexus}, cannot reconstruct the bubbles and hallucinates contents. In contrast, \model\ leverages learned video priors and surface priors to reconstruct the dynamic 3D flows and synthesize plausible novel view videos.

\myparagraph{Interfacial geometry reconstruction.}
We visualize reconstructed interface meshes on real data in Fig.~\ref{fig:real_mesh_static} and on synthetic data in Fig.~\ref{fig:syn_mesh_static}. Quantitative Chamfer distance on synthetic data is reported in Tab.~\ref{tab:numbers}. For all methods, we extract meshes using marching cubes~\citep{lorensen1998marching}. On real data, only \model\ recovers coherent liquid-vapor interface geometry. FluidNexus~\citep{gao2025fluidnexus} and 4DGS~\citep{yang2023gs4d} produce fragmented, incomplete, or even no meshes. On synthetic data, both baselines simply miss most bubbles. In contrast, our \model\ leverages the SDF formulation to enforce geometric priors for reasonable interface extraction.

\myparagraph{3D velocity estimation.}
We show 3D bubble-level velocity estimation results in Fig.~\ref{fig:real_velocity} and Tab.~\ref{tab:numbers}. Since 2DGS~\citep{huang20242d} and 4DGS~\citep{yang2023gs4d} are appearance-focused reconstruction methods, they do not provide velocity estimates. For FluidNexus~\citep{gao2025fluidnexus}, we use its velocity field represented by its physical particles. As shown in the velocity component curves over time, \model\ produces an estimate that closely matches the physically measured velocity, and it is significantly more accurate than all baselines.

\myparagraph{Interfacial velocity visualization.}
We visualize estimated 3D interfacial velocity fields in Fig.~\ref{fig:real_mesh_velocity}. \model\ generates spatially coherent velocity estimates across the reconstructed interface, capturing plausible deformation patterns (such as vertical expansion) and translation during rise. 

\subsection{Ablation Studies}

\begin{figure}[t]
    \centering
    \includegraphics[width=0.91\linewidth]{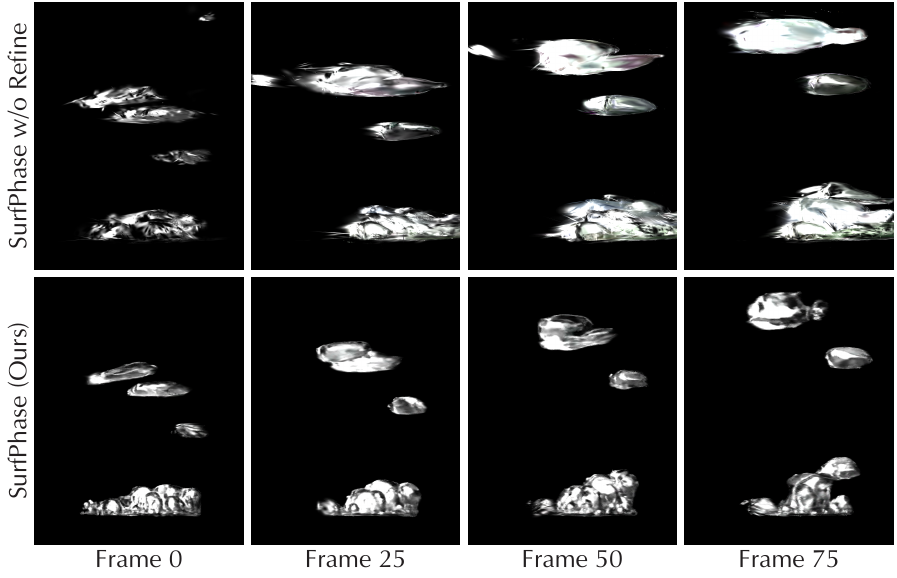}
    \vspace{-1.5mm}
    \caption{Ablation study for video refinement on novel view video synthesis.
    }
    \vspace{-3mm}
    \label{fig:ablation_real_nvs}
\end{figure}

\begin{figure}[t]
    \centering
    \includegraphics[width=0.92\linewidth]{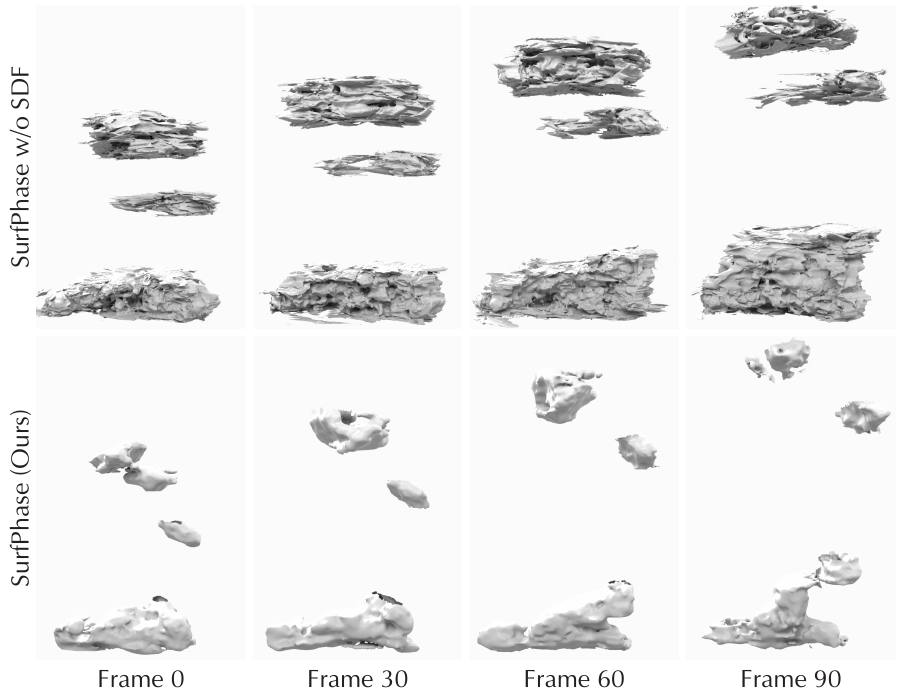}
    \vspace{-1.5mm}
    \caption{Ablation study of geometric surface constraint (\ie, the SDF formulation) on interface reconstruction. 
    }
    \vspace{-3mm}
    \label{fig:ablation_mesh}
\end{figure}

\begin{figure}[t]
    \centering
    \includegraphics[width=0.93\linewidth]{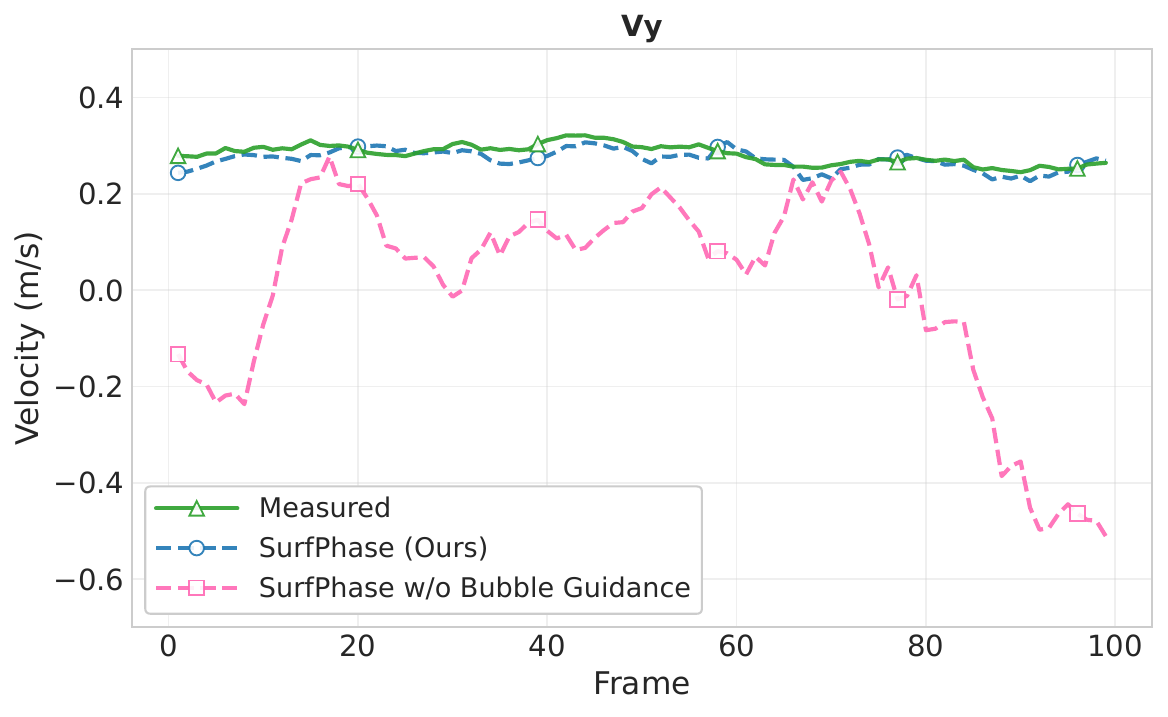}
    \vspace{-1.5mm}
    \caption{Ablation of bubble-guided velocity estimation. 
    }
    \vspace{-3mm}
    \label{fig:abl_vel}
\end{figure}

We conduct ablation experiments on the real dataset to evaluate the core components of \model.

\myparagraph{Video refinement.}
We ablate the video refinement stage by removing the diffusion-based novel-view video synthesis and the subsequent second-stage optimization, denoted as ``\model\ w/o Refine''. As shown in Fig.~\ref{fig:ablation_real_nvs}, without video refinement, the reconstructed 3D appearance contains significant noise and artifacts, particularly in regions not directly visible from the input views. The refined novel-view videos provide learned priors of interface appearance and motion that substantially improve reconstruction quality.

\myparagraph{Geometric surface constraint.}
We ablate the SDF formulation by removing the geometric surface constraint from the Gaussian surfel representation, denoted as ``\model\ w/o SDF''. In this variant, opacity is directly optimized rather than derived from signed distance values. We show interface reconstruction results on real data in Fig.~\ref{fig:ablation_mesh}. Without the SDF formulation, the reconstructed surfels can approximate the visual appearance but fail to form coherent surfaces. Mesh extraction via marching cubes produces overly large, noisy geometry. The SDF constraint encourages surfels to concentrate near the zero-level set, enabling better interface extraction.

\myparagraph{Bubble-guided velocity estimation.}
We ablate the bubble-guided velocity estimation by removing bubble binding and velocity-guided initialization, denoted as ``w/o Bubble Guidance''. This variant performs plain per-frame reconstruction without estimating bubble velocity to initialize surfel positions for subsequent frames. We show velocity estimation results in Fig.~\ref{fig:abl_vel}. Without bubble guidance, the method fails to produce reasonable velocity estimates. This is because differentiable rendering optimization alone does not establish temporal correspondences: surfels tend to remain stationary while adjusting scale and color to match the target appearance, rather than translating to follow the moving interface. The bubble-guided initialization provides physically informed motion priors for velocity estimation.

\section{Conclusion}
We introduced the problem of reconstructing 3D interfacial dynamics in two-phase flows from sparse-view videos, a task that bridges computer vision and experimental fluid mechanics. We proposed \model, which integrates dynamic Gaussian surfels with a signed distance function formulation for geometric consistency and leverages video diffusion priors to compensate for limited camera coverage. Our bubble-guided velocity estimation establishes temporal correspondences that enable metric 3D velocity recovery from as few as two synchronized views. We collected a new dataset of high-speed pool boiling videos for training and evaluation. Experiments demonstrate that \model\ achieves accurate novel view synthesis, interface geometry reconstruction, and velocity estimation, substantially outperforming existing methods.

\section*{Acknowledgments}
The authors gratefully acknowledge funding support from the Office of Naval Research (ONR), with Dr. Mark Spector serving as the program officer, under MURI Grant No. N00014-24-1-2575.

\section*{Impact Statement}
This paper presents work whose goal is to advance 3D reconstruction and velocity estimation for scientific measurement in fluid mechanics. The primary application is experimental research in two-phase flows, enabling non-intrusive measurement of interfacial dynamics. This capability may benefit fundamental research in heat transfer, energy systems, and process engineering. We do not foresee direct negative societal consequences from this work. The datasets collected contain only laboratory footage of physical phenomena with no personally identifiable information.

\bibliography{main}
\bibliographystyle{icml2026}

\newpage
\appendix
\onecolumn
\section{Algorithm.}

We show an algorithm of our proposed \model\ in Alg.~\ref{algo:dyn_bubble_velocity}.

\begin{algorithm}[H]
    \caption{\model: Two-Stage Interfacial Reconstruction with Bubble-Guided Velocity Estimation}
    \label{algo:dyn_bubble_velocity}
    \begin{algorithmic}[1]
    \small

    \REQUIRE Input videos $\{\mathcal{V}^c\}_{c=1}^{C}$, camera poses $\{\pi_c\}_{c=1}^{C}$, novel viewpoints $\{\pi_s\}_{s=1}^{S}$, instance masks $\{\mathcal{M}^c\}_{c=1}^{C}$, video diffusion model $\mathcal{G}$, time step $\Delta t$
    
    \STATE \textbf{Initialize:} Surfels $\{g_i\}_{i=1}^{N_0}$ with attributes $\{\boldsymbol{\mu}_i, \mathbf{s}_i, \mathbf{r}_i, \mathbf{c}_i, f_i\}$

    \STATE
    \STATE \textbf{// Stage 1: Initial Reconstruction from Sparse Views}
    \FOR{$t = 1$ to $T$}
        \STATE Initialize $\boldsymbol{\mu}_{i,t}$ via velocity-guided advection (Eq.~\ref{eq:advection})
        \STATE Optimize surfels by minimizing:
        \[
        \mathcal{L} = \lambda_{\text{app}} \sum_{c=1}^{C} \mathcal{L}_{\text{app}}(I_t^c, \hat{I}_t^c) + \lambda_{\text{geo}} \mathcal{L}_{\text{geo}}
        \]
        \STATE Perform instance binding via multi-view mask projection
        \STATE Update velocities: $\mathbf{v}_{i,t} = (\boldsymbol{\mu}_{i,t} - \boldsymbol{\mu}_{i,t-1}) / \Delta t$
    \ENDFOR

    \STATE
    \STATE \textbf{// Novel-View Video Rendering}
    \FOR{$s = 1$ to $S$}
        \FOR{$t = 1$ to $T$}
            \STATE Render frame: $\tilde{I}_t^s = \texttt{Render}(\pi_s, \{\boldsymbol{\mu}_{i,t}, \mathbf{s}_{i,t}, \mathbf{r}_{i,t}, \mathbf{c}_{i,t}, f_{i,t}\})$
        \ENDFOR
        \STATE Collect rough video: $\tilde{\mathcal{V}}^s = (\tilde{I}_1^s, \ldots, \tilde{I}_T^s)$
    \ENDFOR

    \STATE
    \STATE \textbf{// Video Refinement via Diffusion Model}
    \FOR{$s = 1$ to $S$}
        \STATE Refine with learned priors: $\hat{\mathcal{V}}^s = \mathcal{G}(\tilde{\mathcal{V}}^s; \lambda_{\text{refine}})$
    \ENDFOR

    \STATE
    \STATE \textbf{// Stage 2: Refined Reconstruction with Additional Views}
    \STATE Re-initialize surfels from Stage 1 output at $t=1$
    \FOR{$t = 1$ to $T$}
        \STATE Initialize $\boldsymbol{\mu}_{i,t}$ via velocity-guided advection (Eq.~\ref{eq:advection})
        \STATE Optimize surfels by minimizing:
        \[
        \mathcal{L} = \lambda_{\text{app}} \left( \sum_{c=1}^{C} \mathcal{L}_{\text{app}}(I_t^c, \hat{I}_t^c) + \sum_{s=1}^{S} w_s \mathcal{L}_{\text{app}}(\hat{I}_t^s, \hat{I}_t^{\prime s}) \right) + \lambda_{\text{geo}} \mathcal{L}_{\text{geo}}
        \]
        \STATE \quad where $w_s$ are adaptive weights (higher for views closer to input cameras)
        \STATE Update instance binding and velocities as in Stage 1
    \ENDFOR

    \STATE
    \STATE \textbf{// Velocity Estimation}
    \FOR{each instance $b$}
        \STATE Compute centroid: $\mathbf{c}_{b,t} = \sum_{i: b_i = b} w_i \boldsymbol{\mu}_{i,t} \,/\, \sum_{i: b_i = b} w_i$
        \STATE Estimate velocity: $\mathbf{v}_{b,t} = (\mathbf{c}_{b,t} - \mathbf{c}_{b,t-1}) / \Delta t$
    \ENDFOR

    \ENSURE Reconstructed surfels $\{\{g_i\}_t\}_{t=1}^{T}$, instance velocities $\{\mathbf{v}_{b,t}\}$

    \end{algorithmic}
\end{algorithm}

\end{document}